# Detecting Abusive Albanian

**Erida Nurçe**\*
ITU Copenhagen / Microsoft
nurce.erida@gmail.com

**Jorgel Keci**\*
ITU Copenhagen / UN Development Programme
kecijo@gmail.com

**Leon Derczynski**
ITU Copenhagen
ld@itu.dk

## Abstract

The ever growing usage of social media in the recent years has had a direct impact on the increased presence of hate speech and offensive speech in online platforms. Research on effective detection of such content has mainly focused on English and a few other widespread languages, while the leftover majority fail to have the same work put into them and thus cannot benefit from the steady advancements made in the field. In this paper we present Shaj, an annotated Albanian dataset for hate speech and offensive speech that has been constructed from user-generated content on various social media platforms. Its annotation follows the hierarchical schema introduced in Zampieri et al. (2019b). The dataset is tested using three different classification models, the best of which achieves an F1 score of 0.77 for the identification of offensive language, 0.64 F1 score for the automatic categorization of offensive types and lastly, 0.52 F1 score for the offensive language target identification.

## 1 Introduction

A large number of people are using social media to spread hate and offensive content while using the freedom of speech as a justification for their actions. Simultaneously, studies have shown that an increase in hate speech on social media leads to more hate crimes against minorities in the physical world (Williams et al., 2019).[1] Arcan (2013) explains how hate crime follows hate speech, and the connection between the two can be drawn using the rhetorical stratagem of hate speech described by Slayden et al. (1995).

The growing presence of hate speech on the internet and its connection to hate crimes has emphasised the need for accurately detecting and censoring this type of content while respecting the freedom of speech (Kemp, 2019). A combination of Natural Language Processing (NLP) and Machine Learning techniques have been used for this task. Waseem and Hovy (2016) state the difficulty of detecting hate speech, often due to the lack of sexist, racial or offensive language used in such content. On the other hand, Davidson et al. (2017) discusses how some terms generally regarded as offensive lose their negative connotations when used in certain contexts. Examples mentioned in this paper include African Americans using the term *n*—- in everyday language online (Warner and Hirschberg, 2012), or people quoting explicit rap songs that often include derogatory terms. Furthermore, offensive words are often purposely misspelled or transformed into social media, making them harder to detect.

Hate speech and offensive speech have become more eminent and concerning phenomena in the Albanian speaking subset of the internet. Platforms such as Facebook, Instagram and YouTube have been the main communication medium and source of information for the past few years. Due to the loose regulations and lack of offensive speech detection for the language, the amount of homophobic, sexist and racist comments has exhibited a drastic increase.

Therefore, we aim to contribute to the field by creating the "Spoken Hate in the Albanian Jargon (Shaj)" dataset, a new Albanian dataset annotated using the OffensEval taxonomy (Zampieri et al., 2019a). To benchmark this data, we then experiment with the constructed dataset using various models and see their performances.

---

\*Equal contribution
[1] https://phys.org/news/2019-10-online-speech-crimes-minorities.html

## 2 Background

### 2.1 Hate Speech Regulations

Hate speech is regulated by international conventions such as the Universal Declaration of Human Rights (UDHR) UN-UDHR (2017), the Convention on Human Rights (ECHR) ECHR (1950) and the International Covenant on Civil and Political Rights (ICCPR) which are mentioned in Chetty and Alathur (2018). Countries also have their own legislation regarding this topic (ibid.).

According to Lani (2014), the Constitution of the Republic of Albania does not mention the term 'hate speech'. However, it includes articles regarding the respect towards human rights and freedom, religious co-existence and respect for minorities. Hate speech is regulated in the Criminal Code and it is not only applicable to social media usage. According to (ibid.) the main targets of hate speech in Albania are the LGBT community, minorities eg. Roma, and lastly politicians. Lastly, (ibid.) also states that user-generated comments frequently contain hate speech and not much is being done in terms of regulating or restricting hate speech.

### 2.2 Definitions

Even though the subject of hate speech and offensive language in social media is relatively new, there has been a great number of papers that address the issue all of which have had various definitions of hate speech. As mentioned by Ross et al. (2016) cited in Macavaney et al. (2019), having a concise definition of what hate speech is could positively contribute to the research of hate speech, by making the task of annotation more reliable and easier. This is also reflected in the limitations present in Natural Language Processing (NLP) research Waseem and Hovy (2016). Below, we give some of the most commonly occurring definitions of hate speech.

- Davidson et al. (2017): "Language that is used to expresses hatred towards a targeted group or is intended to be derogatory, to humiliate, or to insult the members of the group."

- Gibert et al. (2018): "Hate speech is a deliberate attack directed towards a specific group of people motivated by aspects of the group's identity."

- Cohen-Almagor (2011) has defined hate speech "as biasmotivated, hostile, malicious speech aimed at a person or a group of people because of some of their actual or perceived innate characteristics"

Hate speech definitions are not only established by researchers, but also from social media giants. Twitter and Facebook have contributed with definitions of their own, all of which are cited in Macavaney et al. (2019).

In many of the articles published, hate speech is closely related to offensive language and even used interchangeably at times. However, in Davidson et al. (2017), a difference is drawn between the terms in hopes of having a more accurate classification. Nonetheless, the work conducted in (ibid.), still lacks a concise distinction between the terms. An example of this misclassification mentioned in (ibid.) deals with song lyrics being mistakenly detected as hate speech.

#### 2.2.1 Criteria

Some of the definitions created throughout the years specify various criteria for classification. Waseem and Hovy (2016) has based the distinction of offensive and not offensive speech on the following criteria:

- Racial or sexist slur
- Attacking minority
- Shows support to problematic hashtags
- Defend xenophobia or sexism etc.

Other related papers, such as Chetty and Alathur (2018), depict a very detailed classification of hate speech detected in social networks. (ibid.) differentiates between two types of hate speech: direct and indirect. When the targeted person or group is immediately impacted by the speech, it is considered to be direct hate speech. Indirect hate speech, on the other hand, indicates that the speech is a starting point of generating more hate. In this paper and also in Gitari et al. (2015), the main types of hate speech are: Gendered, Religious, Racist and Disability.

### 2.3 Annotation Standards

Given that hate speech is widespread and among the most studied subjects of NLP research, it is important to follow predefined practices when creating a new dataset. In Vidgen and Derczynski (2020), the dataset creators put different level of emphasis to the annotation guidelines which leads to a subjective process of annotation. Ibid. also states that it is difficult for annotators to objectively identify irony and intent expression through written. Considering that various factors influence the creation of an abusive language training data corpus, Vidgen and Derczynski (2020) summarize the best practices of such task into the following points:

- Defining the task addressed by the dataset
- Selecting data for abusive language annotation
- Annotating abusive language through clear guidelines
- Documenting methods, data, and annotators

## 2.4 OffensEval Schema

In this work, we follow closely the hierarchy schema developed in the work of Zampieri et al. (2019a) which frames the schema into three subtasks:

- *Offensive Language Detection*
- *Automatic categorization of offensive types*
- *Offensive language target identification*

### 2.4.1 Offensive Language Detection

The goal of the first subtask is to distinguish between offensive and non-offensive language. Offensive content is characterized by the usage of profanity, insults or threats.

Examples of sentences pertaining to these two labels are:

- **Not Offensive (NOT)**: *"edhe 100 klm sot e pergjithmonte puth tezja fort"*

  Translation: *"100 more! (Albanian way of wishing happy birthday). Have a nice time today and always! Kisses from your aunt"*

- **Offensive (OFF)**: *"Shtet pleeeer"*

  Translation: *"Garbage country/government"*.

### 2.4.2 Automatic categorization of offensive types

The goal of the second subtask is to further categorize the data that was annotated as offensive in the first subtask. The two categories of this task aim to distinguish whether the offense is targeted - if the post directly relates to an individual, group or other entity - or untargeted.

Examples of sentences pertaining to these two labels are:

- **Targeted insults (TIN)**: *"Po ti ca je spiun pall"*

  Translation: *"What about you? Are you a spy, you prick?"*

- **Untargeted instults (UNT)**: *"Na hnksh riken"*

  Translation: *"Suck my d*ck"*

### 2.4.3 Offensive language target identification

The last subtask of the annotation schema aims to distinguish the actual target of the offensive text. Zampieri et al. (2019a) identifies three main types of targets: individuals, groups and others. Individual targeting includes insults or threats towards a person, famous or not, named or unnamed. The second label deals with groups of people that share a common characteristic such as ethnicity, gender,(Zeinert et al., 2021) sexual orientation, political affiliation, religious belief, etc. The third label in the subtask corresponds to comments that do not target an individual nor a group, but rather some organization, event, etc.

Examples of sentences pertaining to these three labels are:

- **Individual (IND)**: *"Je shume budalla"*

  Translation: *"You are such an idiot"*

- **Group (GRP)**: *"Ne burg pederastat"*

  Translation: *"Jail for the faggots"*

- **Other (OTH)**: *"Pershendetje joq po cfar i keni keto idiotesira qe postoni se lat nam."*

  Translation: *"Hello JOQ what are these idiotic things that you are posting you are embarrassing yourselves."*

## 2.5 Classifiers

An variety of Machine Learning and Deep Learning models have been used for hate speech classification. Davidson et al. (2017) start their work by running a logistic regression with L1 regularization to reduce data dimensionality. Later on, they evaluate commonly used models such as Naïve Bayes, Random Forest, Logistic Regression and Linear Support Vector Machines (SVMs). The data in this paper is classified into 3 categories: hate speech, offensive language and neither. In their tests, they conclude that the Logistic Regression and SVMs perform slightly better than the other models. For their final model, they use Logistic Regression with

L2 regularization which achieves a precision of 0.91, recall of 0.90 and F1-Score of 0.90. However, they raise the issue of a large amount of hate speech data being misclassified.

In Zenuni et al. (2017) they use Support Vector Machines (SVM) instead of bag-of-words approaches since these produce a high rate of false positives. They try to classify the data as hate speech or not. Their model results in a precision of 0.61, recall of 0.57 and F1-Score of 0.58. Given that the latter model is trained using an Albanian dataset, the results sparked our interest since the work in our paper will try to outscore the aforementioned model.

In Zampieri et al. (2019b), several models were introduced in the competition, all of which were using the Offensive Language Identification Dataset OLID Zampieri et al. (2019a). These models ranged from Logistic Regression to advanced and state of the art deep learning models such as BERT Devlin et al. (2019b) and ELMo Peters et al. (2018). Their work was split into three subtasks. For subtask A, BERT was the dominant model with an F1-Score of 82.9%. For subtask B, the best team used a rule-based approach with a keyword filter based on Twitter language behaviour list. This model produced an F1-Score of 75.5%. For subtask C, BERT was once more the dominant model producing an F1-Score of 66.6%. Even though the best results were achieved from deep learning models, Zampieri et al. (2019b), concludes that machine learning classifiers were among the most used and achieved good results for the task in hand.

## 3 Dataset

### 3.1 Purpose

The main goal behind the creation of our dataset is to improve hate speech and offensive speech detection for less widespread languages. Even though an extensive amount of work and research has been done regarding hate speech, this effort has mainly targeted the English language. This leads to a situation in which the vast majority of the World's languages are still under-resourced in that they have few or no language processing tools and resources (Mossie and Wang, 2018). The Shaj dataset provides a foundation for research regarding hate speech detection for the Albanian language.

### 3.2 Text selection goals and methods

#### 3.2.1 Text selection goals

In order to construct a platform-independent dataset, we have gathered data from multiple accounts in various social media platforms. Our reasoning behind this approach is that the language used online differs depending on the platform's content and user demographics. This way, as mentioned in Bender and Friedman (2018), the dataset will reflect a wholesome portrait of how Albanian language is used throughout platforms including different dialects, writing styles, content types and tone of language.

According to Kemp (2019) the top three social media platforms in Albania are Facebook, Pinterest and Instagram. Due to recent privacy concerns, the bulk extraction of other users' comments on Facebook is no longer possible through their own developer tools (the Graph API [2]). Further on, after taking into consideration that Pinterest is a platform heavily based on images rather than user interaction through the written language, we proceeded with Instagram and YouTube as our main sources of data.

#### 3.2.2 Extraction methods

Comments coming from Instagram were retrieved from two public accounts, namely "@*jetaoshqef* (JOQ)" [3] and "@*lagjia_jone* (LJ)" [4] with the help of an open-source tool named Instaloader [5]. JOQ, the most followed Albanian Instagram account not pertaining to an individual, is an informal news portal whose content encompasses versatile topics and constantly generates high user engagement. On the other hand, LJ is an account dedicated to creating memes, funny and controversial content. Its comment section regularly includes examples of offensive speech and hate speech, often expressed in slang that you would not encounter in formal media outlets. Data obtained from each of the sources contains the comments of the 100 most recent posts at the time of retrieval (October 2019).

Comments coming from Youtube (YT) were retrieved also from two channels with the help of Google's Youtube Data API [6]. According to So-cialbakers [7], RTV Klan [8] is the second most-viewed Youtube (YT) channel in Albania. We, therefore, extracted comments from some of their most commented videos. Furthermore, comments were also collected from episodes of one of the TV shows with the highest audience numbers of this TV channel, which are posted on another account, namely "Ermal Mamaqi"(the host of the show) [9].

---

[2]https://developers.facebook.com/docs/graph-api/reference/v5.0/object/comments
[3]https://www.instagram.com/joqalbania/
[4]https://www.instagram.com/lagjia_jone/
[5]https://instaloader.github.io/
[6]https://developers.google.com/youtube/v3/getting-started

### 3.3 Ethics and Privacy

When handling this kind of data, it is important to ensure the anonymity of the speaker. To conform with the General Data Protection Regulations in Europe (GDPR)[10], we have taken some precautions when dealing with speaker's identity. Firstly, every username has been replaced with the *@USER* tag to avoid direct addressing of the speakers. Furthermore, we were cautious to not include sensitive user data, such as addresses, phone numbers, emails, etc. in our data. This approach is also used in other works such as Sigurbergsson and Derczynski (2020), Zampieri et al. (2019b) and Zampieri et al. (2019a).

### 3.4 Annotation Procedure

During the annotation procedure, 4 different annotators participated in the process including the authors of this paper. In order to mitigate annotator bias (Waseem, 2016), the annotators first started a process of annotating a portion of the same comments individually, later comparing the given labels and having discussions about their thought process. The final dataset annotation was again reviewed from the authors of this paper.

### 3.5 Annotator demographic

Annotator demographics is an essential part of the annotated corpus creation, as emphasised in Bender and Friedman (2018). The annotators of Shaj ranged for all the specifications mentioned in (.ibid) except their native language, which was Albanian. Male and female annotators between the ages of 22 and 25 years old participated in the process. They come from different cities of Albania and belong to different religious groups (Muslim, Orthodox, Catholic, Bektashi). Experience wise, all of the annotators were involved in such a process for the first time and based their decisions on the annotation schema provided to them combined with their personal experience with common uses of language in social media.

### 3.6 Examples of data corpus

During the annotation process, a variety comments were present, with some of them raising a challenge regarding their appropriate labelling. The main issue we encountered relates to the comment's context. It was at times uncertain whether some phrases where being used as insults or simply stating facts relating to the image of the post or a sequence of a video. Such an example can be raised when comments included the word "*kungull*" which translates to "*pumpkin*" in English. This term is often used in the Albanian language to describe someone who is stupid (their head is hollow as a pumpkin 🎃, thus missing a brain). Without the context of the post, it was difficult to differentiate whether the word was being used literally to describe the vegetable or as a metaphor to offend someone. In this case, the original post was traced and checked in order to provide the most appropriate label.

Another issue we faced during this process, is the multiple meanings of the words used in the comments. A lot of words in the Albanian language can be translated into insults and profanity given the complete sentence structure. When dealing with these kind of comments, the annotators, who have been used to the social media jargon and phrases, raised discussions and opinions that lead to a label as a final verdict from the authors. Examples for this case can be the usage of "*ME PLASI ADMIRALIIIII*" which roughly translates to "*My admiral does not care*, however the *admiral* word is also used in social media to refer to a male's genital organ.

During the annotation process there were also clear cases that helped us give a correct label to the data corpus. The usage of the personal pronouns such as you, him/her, they, them etc., along with the use of profane words/phrase helped us identify the target of the offensive and hate speech. Such an example for targeted insults is *"@USER ti je kokrra kurves si se ke iden per ket gje thjesht do si njerzit te ndjekin ty"*, translated to *"@USER you are a slut you have no idea about this thing, you only want people's attention"*. An example for NOT label is *"@USER un them se nuk eshte e vertet"*, which translates to *"@USER I do not think this is true."*.

---

[7]www.socialbakers.com/statistics/youtube/channels/albania
[8]https://www.youtube.com/user/televizioniKLAN
[9]https://www.youtube.com/user/ErmalMamaqiOfficial
[10]https://gdpr-info.eu/

## 3.7 Final dataset

Table 1 displays information about the final Shaj [11] dataset according to their sources and distributed labels. The final size of the dataset was 11874 comments.

From table 1 we observe that the dataset is skewed towards not offensive (NOT) content that makes up 87% of the whole corpus. However, as seen in works of Sigurbergsson and Derczynski (2020), Zampieri et al. (2019a) and Çöltekin (2020), this skew is evident in many datasets that portray real life user-generated content on social media platforms. This argument is also re-enforced when we have tried to cut down some repetitive comments or comments that tag one or multiple people. Such instances have not been removed completely to maintain a true representation of real-life content.

While gathering the comments from different platforms, we noticed the importance of the emoticons used extensively in the daily online language. These emoticons can express different emotions or opinions ranging from a happy smiley face to an insult. A concrete example from our dataset would be the sentence "*ky duhet #EMOJI_FIRE*". Without the emoticon the sentence translates to "*you need this guy*". However with the inclusion of the emoticon the meaning of the sentence changes to "*you should burn this guy*". We can see that social media text comprehension often relies on emoticon usage and therefore, decided to keep the emoticons as an important part of the dataset. This choice is also seen in the work of Gitari et al. (2015), whereas Albadi et al. (2018), Zenuni et al. (2017) choose to remove emoticons as part of their data cleaning process. Orts (2019) on the other hand follows an intermediate approach by translating emojis into words.

| Label/Source | JOQ | LJ | YT | Total |
|---|---|---|---|---|
| NOT | 9203 | 902 | 201 | 10306 |
| OFF, UNT | 384 | 55 | 9 | 448 |
| OFF, TIN, IND | 644 | 30 | 64 | 738 |
| OFF, TIN, GRP | 220 | 7 | 7 | 234 |
| OFF, TIN, OTH | 134 | 7 | 7 | 148 |
| **TOTAL** | 10585 | 1001 | 288 | 11874 |

Table 1: Statistics for each datasource of the Albanian dataset

[11] Shaj is a backronym for the Albanian verb that translates to offend, insult, curse, swear, etc.

## 4 Automatically detecting abuse in Albanian

### 4.1 Model introduction

We choose BiLSTM as the main model to perform abusive speech detection for Albanian language. The model is firstly comprised of the embedding layer which will use vectors representation of words in our corpus. These vectors come from the FastText library [12]. For words that do not have a vector representation from FastText, but appear in our corpus, we construct the same sized random vector as FastText following a normal distribution. The next layers of the model consist of the bidirectional long short-term memory (BiLSTM) layer, a fully connected hidden layer and an output layer. The bidirectional LSTMs consists of 100 nodes each. The activation function used in our case is *RELU*, the output activation function used in *softmax*. The optimizer used is *adam* Kingma and Ba (2017) and the loss is calculated with *sparse_categorical_crossentropy*.

### 4.2 Data split, tuning and metrics

#### 4.2.1 Data split

Having described the corpus size in Section 3.7, we use a 80-20 split ratio between training and testing the model. This 80-20 is kept for each of the labels we have. Table 2 describes the details of this split.

| Label | Training | Testing |
|---|---|---|
| NOT | 8245 | 2061 |
| OFF, UNT | 358 | 90 |
| OFF, TIN, IND | 590 | 148 |
| OFF, TIN, GRP | 186 | 48 |
| OFF, TIN, OTH | 118 | 30 |

Table 2: Statistics for each datasource of the Albanian dataset

#### 4.2.2 Model Tuning

Table 3 displays the parameters that were used to conduct the experiments for the BiLSTM model. As less and less data belongs to the more specific subtasks expressed in Section 2.4, we have increased the number of epochs to allow the model to train more. The tuning process was established by conducting many experiments with different parameters and observing when the model performed

[12] https://fasttext.cc/

best. During these experiments, we present the parameters that have not shown model to overfit.

| Subtask | Epochs | Opt | LR | Batch |
|---------|--------|------|-------|-------|
| A | 20 | adam | 0.001 | 128 |
| B | 20 | adam | 0.001 | 128 |
| C | 50 | adam | 0.001 | 128 |

Table 3: Parameter tuning. Note that LR=Learning Rate and Opt=Optimizer. All subtasks train the initial weight of the embedding layer

### 4.2.3 Metrics

We choose F1 score as the main metric for our model's performance. F1 score is calculated as $F_1 = \frac{2 \cdot precision \cdot recall}{precision + recall}$ and gives a better overall understanding of the model's results since it evaluates the scores of each class independently and calculate an unweighted average of these. Its benefits over metrics such as accuracy would be that when having the model classifying a lot more samples of one class would produce a high accuracy, but a low macro average F1 score. Thus, we use the macro averaged F1 score to see if our models have been over-fitting.

### 4.3 Baseline

In order to compare our model scores, we introduce Naïve Bayes as a baseline model. The algorithm used in Naïve Bayes has its synthesis from the Bayes Theorem. It calculates the probability of a class *c* given some initial information about a feature vector *x*. The Naïve part of the name comes from the assumption that is being made when using the theorem that every pair of features given a value of the class variable are conditional independent. In this paper, we use the Gaussian distribution and algorithm used in sklearn library [13] to implement such baseline.

### 4.4 BERT

Deep-learning models are not the only ones used when detecting offensive speech. The transformer approach has proven to have high results with the task in hand and therefore we include BERT Devlin et al. (2019a) as one of the models that uses this approach. We use BERT base model consisting of 12 layers, 768 hidden size and 12 self attention heads. The Huggingface library [14] includes a variety of BERT models, but we have chosen to work with the base case of *bert-base-multilingual-uncased*.

---

[13] https://scikit-learn.org/stable/modules/naive_bayes.html
[14] https://github.com/huggingface/transformers

## 5 Analysis and Discussion

In this section we provide the results obtained from the three different models described in **Section 4** according to the subtasks described in **Section 3**.

### 5.1 Offensive Language Detection

In table 4, we share the results achieved for the first subtask, according to the F1 score and observe that BERT achieves the best performance with a 0.77 F1 Score. The results are then followed by the BiLSTM model with a score of 0.70. Comparing our results with similar papers that propose new datasets such as Sigurbergsson and Derczynski (2020) or Çöltekin (2020), we obtain comparable results, indicating an appropriate dataset for Albanian hate speech detection.

| Model | Dataset | Precision | Recall | F1 |
|-------|---------|-----------|--------|------|
| NB | Shaj | 0.51 | 0.50 | 0.41 |
| BiLSTM | Shaj | 0.67 | 0.74 | 0.70 |
| BERT | Shaj | 0.74 | 0.81 | 0.77 |

Table 4: Results for subtask A using Shaj dataset

### 5.2 Automatic categorization of offensive types

In Table 5, we share the results achieved from the second subtask according to the F1 score. The best scoring results are again achieved from BERT with a 0.64 F1 score. In these experiments, we see a drop in performance in BiLSTM and BERT. This is due to the amount of data used for the second subtask.

| Model | Dataset | Precision | Recall | F1 |
|-------|---------|-----------|--------|------|
| NB | Shaj | 0.51 | 0.51 | 0.48 |
| BiLSTM | Shaj | 0.62 | 0.64 | 0.63 |
| BERT | Shaj | 0.63 | 0.67 | 0.64 |

Table 5: Results for subtask B using Shaj dataset

### 5.3 Offensive language target identification

In Table 6, we share the results achieved from the third subtask according to the F1 score. Due to the small amount of data that the models are run on, we see a significant drop from the first two subtasks. In this subtask, we also increase the number of epochs so that the model can train longer, keeping in mind not to overfit it. Again, we observe that BERT is the best scoring model according the F1 score.

| Model | Dataset | Precision | Recall | F1 |
|---|---|---|---|---|
| NB | Shaj | 0.29 | 0.56 | 0.25 |
| BiLSTM | Shaj | 0.51 | 0.54 | 0.51 |
| BERT | Shaj | 0.51 | 0.55 | 0.52 |

Table 6: Results for subtask C using Shaj dataset

During the different experiments performed in this paper, we see a gradual decrease in results as with each subtask. This is because less data that is being fed to each of the models. An additional factor for the third subtask results is the complex concept of the OTH target, which is often assigned when none of the two other targets types are appropriate rather than having a very clear definition of its own.

A comparison between BERT and BiLSTM results in subtask B and C, suggests that BiLSTM and BERT have little difference when dealing with smaller datasets. However, judging by the results or subtask A, the pre-trained and fine-tuning approach established with BERT, gives insights that the transformer approach is better prepared with the language used. The difference between the approach of the models where BERT does not make use of the word vectors, help its performance when learning the contextual relationship in a text.

The parameters used when conducting the experiments also play a role in the results of the subtasks. The experiments shown in this paper, followed a fine-tuning process of the parameters, ensuring that the model would not overfit during training, by also constantly validating the training results against a smaller portion of validation data.

### 5.4 Discussions

The experiments performed in this paper helped us identify a few interesting points about our dataset and also brought into light some of the common pitfalls of hate speech detection.

The models helped us identify mis-annotated data in our corpus. After running the experiments, we observed entries in our corpus that at a second glance were not correctly annotated. A point made in Vidgen and Derczynski (2020) and also in Section 3.5 has been identified that annotators demographic features and social upbringing have influenced the annotation label. An example of this issue is:

- *ku ka magji qe i mer ato persiper m* translated to "not even magic can take care of those"

Failures of the model on identifying the intent, irony or even sarcasm by the model were depicted in other results.

- *tamam parkim bjondeje* translated to "typical blondie parking"

- *bjonde do kete qene me siguri* translated to "she must have been a blonde for sure"

These problems are known in NLP, as stated in Vidgen and Derczynski (2020).

The issues discussed in the previous paragraphs, however, are a great example of the diversity of expression that characterises the Albanian language. Sentence structure, words with multiple meanings, context, irony and sarcasm make hate speech detection even more difficult and our contribution with the dataset more valuable towards the improvement of hate speech detection in social media.

## 6 Conclusion

Hate speech in social media is a concerning phenomenon that has gained attention in the recent years. Effective detection systems for all languages are required to minimise such content.

We have presented Shaj, the first publicly available Albanian dataset for hate speech detection. The dataset of ∼12000 entries is comprised of Instagram and YouTube comments and annotated using the OffensEval schema.

We have experimented with various models such as Naïve Bayes as our baseline, BiLSTM and BERT to see how well hate speech can be detected using Shaj. While our baseline has an F1 score of 0.41 in detecting offensive speech, we achieve an F1 score of 0.77 using BERT.

The outcomes of our experiments show that while detecting hate speech is difficult, in part due to irony, sarcasm, or lack of context, it is still possible to achieve useful results on new languages with the help of well-constructed and annotated datasets combined with advanced machine learning and deep learning models.

Shaj is available under CC-BY 4.0 at https://doi.org/10.6084/m9.figshare.19333298.